# EXTENDING A MODEL FOR ONTOLOGY-BASED ARABIC-ENGLISH MACHINE TRANSLATION (NAN)

Neama Abdulaziz Dahan[1] and Fadl Mutaher Ba-Alwi[2]

[1]Department of Computer Science, FCIT, Sana'a University, Sana'a, Yemen
[2] Department of Information Systems, FCIT, Sana'a University, Sana'a, Yemen

## ABSTRACT

*The acceleration in telecommunication needs leads to many groups of research, especially in communication facilitating and Machine Translation fields. While people contact with others having different languages and cultures, they need to have instant translations. However, the available instant translators are still providing somewhat bad Arabic-English Translations, for instance when translating books or articles, the meaning is not totally accurate. Therefore, using the semantic web techniques to deal with the homographs and homonyms semantically, the aim of this research is to extend a model for the ontology-based Arabic-English Machine Translation, named NAN, which simulate the human way in translation. The experimental results show that NAN translation is approximately more similar to the Human Translation than the other instant translators. The resulted translation will help getting the translated texts in the target language somewhat correctly and semantically more similar to human translations for the Non-Arabic Natives and the Non-English natives.*

## 1. INTRODUCTION

Misunderstanding the meaning of a text in articles, contracts or even books may lead to many problems in a way or another. Thus, we need to translate the main idea instead of the statistical recorded meaning of the words to avoid such problems. Translation is the conversion of the meaning of a text from a language to another language in a precise manner which requires the understanding of the original text and then expressing the content and forms in the other language [1]. Human translators are still the best to translate such texts [2] because of the provided precision and semantic correction of the meaning. However, their translation is so expensive, cannot be available all the times for every single word we may need [2] and need time to be ready without translation errors. In addition, the lack of the translator experience will lead to a bad translation [3].

From the above points, the machine translation scientists started the instant translation [4] which solves the cost and translation errors according to the availability of the telecommunication techniques. The instant translator is an essential means of support for  the connectivity of people in their redundant communications [4]. As a field of the Natural Language Processing, to have a Machine Translation (MT) you can merge many fields of science to complete the translation process [5], such as translation itself, algorithms analysis, web technologies especially the semantic web and artificial intelligence fields especially the machine learning. MT techniques vary and can be classified into three main categories: Statistical Machine Translation (SMT), Machine Translation using Semantic Web (SWMT), and Neural Machine Translation (NMT) [2]. SWMT is an MT system that uses the semantic web techniques to provide the final translations [10]. The ontology files are used in this type of MTs as a knowledge-base to store and retrieve the requested translations. One SWMT may have many ontology files, one per a specific context from a general pre-identified context [2]. However, only one ontology file can be used at a time. In addition, the wrong ontology specification will lead to translation error(s) [10].

As NMTs, SMTs or MTs uses the English language as a pivot language [6], the resulting translation can be correct as long as one side is the English language. However, this technique is not perfect all the times, especially if the other language has different characteristics like the Arabic language. Although the Arabic language is one of the UN formal languages [7], the MT translation from or to the Arabic language using any MT category has not been in the same level of the human translation because the MT either need to store all the potential translations as corpus, such as the united nation corpus, or will provide a literal translation as dictionaries, such as the Oxford Arabic-English dictionary. The problem is not only having correct English-Arabic MTs only, but also the flexibility and the morphology of the Arabic language [8, 9] which leads to the whole translation errors. The Arabic structure is as flexible as having a sentence whose object is written before its verb and subject or to having a complement before the main verb or even the idea of the sentence. The translator must rearrange the target sentence before sending it to the receiver which is the reason of many problems while using the MT. In addition, the Arabic languages according to many dictionaries and lexicons has rich morphological rules. However, this is leading to an ambiguity while translating, because many different meanings can be understood from one single word. The words can be homonym which means that two or more words share the same sound but differ in meaning and homograph which means that two or more words share the same spelling but differ in meaning. According to Moussallem, et al, [10] these types of words in the English language are either nouns or verbs at most and not all the English words may be homonyms, whereas the Arabic language has many homonyms and homographs at the same time and they are not only the verbs or nouns [9] because the prepositions in the Arabic languages are all homographs.

As an example that can show the translation errors in google and Bing instant translator is " يقفون أمامها متأملين أبعادها" which means that "**to stand contemplating its dimensions**", according to the center of Translation and Language Training (CTLT).   Google's translation is "They stand in front of her, hoping for her dimensions." And Bing's is "They stand in front of her, hoping to keep her away." In the reverse translation, Google's is "الوقوف متفكرين في أبعادها" Which means that "Thoughtful standing in its dimensions". Bing's is "للوقوف علي التفكير في ابعاده" Which means that "Ali, for standing and thinking in its dimensions." Google and Bing choose the most statistical recorded meaning to be written as a translation for the word which leads to the misunderstanding of the final resulted text. It might be acceptable in the English language according to have many synonyms for some words, but in the Arabic language which is richer, it may not be acceptable because each word has its unique meaning which is slightly different from the meaning of its recorded synonyms. The intended MT is to provide a translation that is going to be as correct as CTLT's. Sentences need to be divided using a parser to have its real stem and POS. In the Arabic language, the word may have many parts which lead usually to translation errors. In addition, replacing word by word leaves the sentence in the structure of the source text. Replacing phrase by phrase leaves some bad translation of the homonyms of the sentence. Having a parser that divides the sentences into its real words, replacing the words with their correct meanings and then reordering the words according to the structure of the target language is the human way in translation and hence the adopted way in NAN. Regardless the time it takes, this way gives more correct translation than Google's and Bing's.

As a result, to provide a somewhat correct translation that is somewhat similar to the human's with less error rate, the translation must depend on the meanings before using any MT techniques. The English-Portuguese Judge Method (JM) model [10], as an example of the semantic-based MT model, is used to disambiguate the homographs to provide a correct translation. JM is not the effective solution for the Arabic-English language, because it was built for the English-Portuguese translations only where this couple have the same structural features and do no need to rearrange the resulted text. Dahan, et al [9] had extended the JM model before, but their model wasn't approved and failed on the first try to provide a translation because the sentence still needs to be rearranged before sending the final translation. Thus, we are going to re-extend the JM model to serve the Arabic-English MT and to solve Dahan's model translation errors. The new extended model will be named NAN to express that it will serve the non-native speakers, because it will serve the Arabic and the English foreign students, traders, and researchers. Then, to evaluate NAN, a sample dataset is supposed to test the model while the trial and error approach because all the accredited benchmarks cannot show the difference in the resulted translations. In this way, a semantic similarity function will be used to evaluate the resulted translation by NAN, Google, and Bing in comparison with the human translation. Each word inside the single document is converted to a vector. The weight of a vector is computed according to the frequency of its occurrence in the document. It is chosen because the traditional keyword matching fails while comparing the semantically or lexically related documents and we actually pay for the correctness than the time or speed as it is know that this type of MT is considered the slower. The rest of this research paper is going to be in the following order: II. Literature Review, III. The Proposed Model (NAN), IV. The Model Evaluation, V. Conclusion.

## 2. RELATED WORK

In In Machine translation field, the instant translator is a term appeared in F. G. Shinn in 1898 while converting the feelings to piano music [11]. In 1953 it was discussed in terms of translation for words whose originality is not known. In that research, the problem of words' ambiguity was discussed and the problem was caused by the disability to define the correct translation of an explored term [12]. In 1966, it appeared again, but this time it for speech translation and at the same time, the language grams appeared [13].

As mentioned above, the MT has different categories. SMT is an MT system that is based on the machine learning methods, such as Bayesian [14] and clustering algorithms [15]. Its techniques include word based, phrase based, syntax-based and hierarchal phrase based translations [16]. It is good as a dictionary for words, idioms or famous clauses [2]. However, its translation is not always correct, because the final translation depends on a high rate that is generated by the clustering algorithms according to the more accepted translations [10]. SWMT is an MT system that uses the semantic web techniques to provide the final translations [10]. The ontology files are used in this type of MTs as a knowledge-base to store and retrieve the requested translations. The single SWMT may have many ontology files; each one for a specific context from a general pre-identified context [2]. However, just one ontology file can be used at a time. In addition, the wrong ontology specification will lead to translation error(s) [10]. Based on the Artificial Neural Network (ANN), NMT is used to make the MT learn concurrently to increase the performance of the MT through both the encoder and decoder which are needed as the two steps of the Recurrent Neural Network (RNN) [17-20].

As examples of the SWMTs, there are different types of research. One of them has built a Chinese-English ontology that can be used to generate a somewhat precise translation derived from the ontology [21]. Another translation was using the ontology to do the syntactic and the semantic analysis in order to do an English-Korean MT [22]. In addition, the ontology was used by Moussallem, et al, in [10] as a core of the semantic analysis phase in the dialogue system model that provides an English-Portuguese translation. In contrast, the ontology file was used as a corpus file to be a dictionary while building a parser or to enhance the Neural Machine Translation (NMT). According to the best of our knowledge, there's no machine that can provide the Arabic-English translation till this moment [23], especially what is based on the semantic web, and all the translators found still suffer from the translation errors because of the ambiguity of the homographs [10, 24]. All the available machines we found while we researched were either parsers, statistical MTs, NMTs or DLMT, or MT based on the Semantic Web for many languages, except for the Arabic Language. In this research, two ontologies may be included inside the machine, and there will be an interface to receipt the source language and to provide the target language.

Moussallem, et al [10] developed a model for Portuguese-English MT to be used in the dialogue systems. As we mentioned earlier, specifying the homographs is needed to disambiguate the homographs in order to know what the best meaning is to be chosen using the main context of the sentence. If the morphological analyzer finds a homograph, it will send it to the semantic analyzer with the sentence context to be looked for inside the ontology file. The semantic analyzer is needed to choose the best meaning of the homograph according to the sentence context. The Judge Method is effective according to the coherent architecture of the semantic analyzer to make analysis for the sentence more than once. The meaning of each homograph will be replaced by in the source sentence. After that, the model is to use the statistical analysis to get the final translation that the machine will send to the target, using an instant translator to provide some speed for the machine [10]. The ambiguity problem is still a problem in the Arabic language, which is probably the main reason for the absence of complete and efficient translation machines in the state of the art. Some of the researchers deal with the parser and the morphological analyzer as being the same tool, because they have the same mechanism [10, 25]. However, other researchers deal with the morphological analyzer as being one of the components needed for building the parser [26, 27]. In this research, we will deal with the parser as it has a morphological analyzer as an inner component that is needed to be used with the English language as it is without the other components of the parser.

These days, there are many types of research that are focusing on the Arabic-English Machine-Translation or something related. Elaraby et al [28], discuss the gender agreement violations as the source of the translation errors. In their opinion, it is not a problem while translating among Arabic, Turkish and Japanese, because a person has many words that point to his/her gender from his/her speech. However, the problem will start appearing when translating between Arabic and English, because there are so few words that may be used to know the gender of the person who is speaking or absent in the speech. Alkhatib et al [29], classified the Arabic-English MTs' problems into Lexical, syntactic and semantic problems. The length and punctuation of the Arabic sentences are some of the real challenges too. Its flexibility makes it a rich language but also makes it one of the languages that are hard to have an MT. Therefore, in our opinion, this is why the Arabic-English MT still has many errors and challenges. Pal [30] stated that a hybrid framework for MT can be better than other types of frameworks and can give a better translation.

## 3. THE PROPOSED MODEL (NAN)

As shown in figure 1, NAN, the proposed model was extended and enhanced to be used for the Arabic-English Machine-Translation, based on the ontology. Although NAN adopted the Method Judge as the heart of the MT, it didn't use the inner components in the same order besides the extended components. To have a model that serves the Arabic-English MT, we used the trial and error method to the extension process. The bold components are what was extended and the rest are the main JM components. They were added to ensure some correction of the sentence grammar and structure, and some correction of the words meanings which the MT writes during the translation. Because of that, in each step, we either added a new component or rearranged the previous components to have the final some correct result.

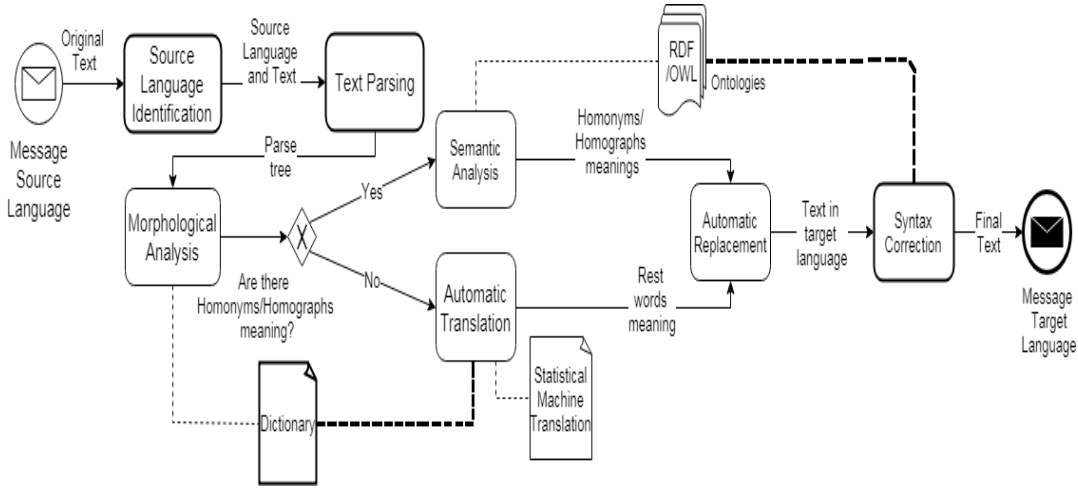

Figure 1: The Non-Arabic/English Natives (NAN) model

## 3.1 Identifying the extension Process:

To adopt the human way in translation, we had to deal with each one of both languages according to their particular characteristics, due to the difference between them. Therefore, the model should specify the source language before doing any process in order to provide the final translation. According to the source language specification, the model should choose the language-oriented processes, either the Arabic or the English language to have special parser, special morphology, and special correction algorithm. In addition, the language-oriented processes is to identify the text context to have a special ontology file. The source language sent the source text to the language parser, which is added to identify the text parts-of-speech (POS) taggers. Both languages should have a parser to identify the homographs. However, the parsers should be different according to the differentiation in the language characteristics. The resulted parsed text of the parse tree would after that be sent to the morphological analyzer to identify wether the single word in the text has a single or multiple meanings. According to this division, both results will be taken and processed separately. The single-meaning words would be sent to the automatic translator to be converted to the target language using either a simple dictionary any other SMT. The multiple-meaning words would be sent to the semantic analyzer to read their meanings from the related ontology file which was identified previously using the context of the source text. These two results would then be sent to the automatic replacement to replace the source texts with their equivalent meanings in the target language. The problem was that the result of the auto-replaced text is meaningless and has incorrect grammar although its literal meaning is correct. Actually at this step the sentence is still in the structure of the source text. Therefore, the model needs to implement Simha's correction Algorithm [31] to rearrange the resulted text before sending the user the final translation. As a result, the resulted text was reasonable and acceptable.

### 3.1.1 Getting the Dataset:

The previous researchers have developed some benchmarks, such as the United Nations six parallel corpus. However, they are not convenient to prove the accuracy of this model against the instant translators because these benchmarks are stored in the knowledge base of the instant translators, therefore the accuracy of the instant translator is not different from the human translation, but they are the same. The problem which this research is going to solve appears while translating non-benchmark texts from or to the Arabic-language, such as book texts, contracts or research articles. As a result, we prepared a new dataset. We chose four paragraphs in the Arabic language and four other paragraphs in the English language according to their bad translation in Google and Bing instant translators. Then we asked the Center of Translation and Language Training (CTLT) in Sana'a University - *as they are considered as the center#1 which is accredited to translate such documents* - to provide us their correct and accurate translation to be accredited as the human translation. The resulted dataset contains 16 paragraphs. Each paragraph has from 1 to 5 sentences and each sentence has from 1 to more than 7 phrases. They were all the training and the test data. Its size was calculated using equation 1 and it was equal to 288.7247824. Although it seems a small dataset, it contains many different structure cases and multiple homonyms and homographs which show a real difference between human translation and the instant translator, to solve the problems of the used benchmark mentioned above.

$$Size(m) = \sqrt{\sum_{r \in m} length(r)^2}$$

Equation 1: Size of the dataset

### 3.1.2 Developing the Ontology files:

Since the model is providing a translation based on the ontology, we developed four different ontology files, from scratch and those will be used to make the semantic analysis. We didn't use pre-found ontology files because of two reasons: the first reason is that the number of the new classes and relations that we will add to those ontologies are almost everything needed while translation. The second one is that the comprehensive ontology files are very huge, which causes being delayed for too much time before the result can be shown, which, in some cases, can be a long time; up to one day with no response. It may cause another problem when the term itself can have more than one meaning in the same ontology file which contradicts with the original model hypothesis which is assigning one meaning per a language for each class or relation defined inside the ontology file. The dataset is eight paragraphs, each two are talking about the same topic and thus the homonym words have only one meaning in one context. The ontology files were developed from the dataset itself. The homographs are recorded in with their equivalent meaning in the Arabic and the English languages. We used Protégé to develop the ontology files and the corresponding java classes for each class in ontology.

### 3.1.3 NAN trails and error:

To test the model, NAN-MT had a sample implementation in C# code and each component mentioned in figure1 has been added to the C# windows form application as a service/component resue or a simulation for its mechanism if the component is not really found yet or if it is only working online. While each component needs an input in a specific format, we use the output of a component to be the input for another. The reused services were: RestSharp is an online web service that was used to detect the text language automatically, Microsoft ATKS as the Arabic parser, and Google API translator as the SMT. These services were also replaced by offline code that simulates their processes or at least offers their functionalities. RestSharp was replaced by two radio buttons to make the user identify the source language manually. The parse trees of ATKS was recorded in XML file to be used offline. Google API was replaced by XML dictionary file.The reused component is Stanford CoreNLP as the English parser. As the researchers couldn't find a source offline code for the morphological analyzer or the dictionary, they simulated their works using a sample user defined dictionary in XML files. The Ontology files were created from scratch to have the needed terms. The semantic analyzer was coded by the researchers. The Correction algorithms was mentioned in Simha research [31], to rearrange the target sentence according to one acceptable structure in the target language, without any indication to its implementation. Therefore the researchers has implement its work partially. All the XML files and the ontology files were built for the research purposes.

According to that way in the implementation, the translation errors may be caused according to errors in the parse trees, errors in specifying the ontology files or lacking in its recorded terms, or errors in the correction algorithm implementation. The parse tree identifies the real parts of the sentence according to the sentence grammar and words stems. The need of its usage is to specify the different structures and properties of the sentence language. The results of the parse tree are the text words, their POS taggers.

### 3.1.4 NAN trails and error tests Examples:

Using the previous model, JM, the translation resulted in errors. For example, the English sentence: "**The point where images were projected on a screen in a darkened theater**" means in Arabic, according to CTLT: "النقطة التي فيها يتم عرض الصور على شاشة في مسرح مظلم". According to the JM translation, its meaning will be "شاشة a ال النقطة التي فيها الصور يتم عرض على مظلم مسرح a في", which means "The point where in it the images had viewed on a screen in a theater darkened". Making the translation after the replacement caused some problems for some words like the "a" between two homographs and homonyms. Thus, we rearranged the JM components and implement a sentence reordering algorithm to provide a correct structure for the final sentence. The morphological analyzer was not found separately and had limited functions to identify the POS taggers. It is used to calculate the number of the meanings of the words. Therefore, we added a parser component before it to make it have the full functionality. When we tried to do the same with the Arabic language, a very bad translation was provided. For example, the Arabic sentence: "وبعد الاسلام أصبحت شبام مدينة عامرة" means in English, according to CTLT: "**After Islam, Shibam became a populated city**". The parser was bad to identify the real and correct POS taggers. The NNP is just NN and the VBD is NN and so on. The resulted JM translation was: "**And after Islam became Shibam city populated**". In addition, the compound verbs or verb-tense additions were being searched inside the ontology, too to avoid the translation errors as much as possible. Therefore, we used another parser for the Arabic language and implemented the correction algorithm for two-sided translation. The final Nan model is shown above in figure 1. And the example and its trace is shown in table 1.

$$\cos \theta = \frac{d1.d2}{|d1|.|d2|}$$

Equation 2: the Cosine Similarity Equation

As the dataset is small and semantic-based, the cosine similarity is supposed to evaluate the model and its output. All the dataset paragraphs are converted to vectors. Cosine similarity can be used to show the similar lexical terms within the compared documents. Therefore, each word in each paragraph will be considered as a vector. The comparison will be among Google, Bing, NAN and CTLT translations. The cosine similarity equation #1 can be explained as follows: Where $\cos(\theta)$ is used to calculate the angle between two sentences, paragraphs or documents vectors, d1 is the first document which will always be considered as CTLT translation, and d2 is the second document which will be considered as google, NAN or Bing translation. |d1| is the length of the first document and |d2| is the length of the second document. Each paragraph is dealt with as a document. The tables 1 and 2 show the registered values for each one.

Table 1: NAN trails and error test example

| Process | English Source | Arabic Source |
|---|---|---|
| **Language Identification** | The point where images were projected on a screen in a darkened theatre: English Source, Cinema Context | وبعد الاسلام أصبحت شبام مدينة عامرة: Arabic Source, Shibam Context Means: After Islam, Shibam became a populated city. |
| **Parser** | ROOT(S (NP (DT the) (NN point))(SBAR (WHADVP (WRB where)) (S(NP (NNS images))(VP (VBD were)(VP (VBN projected) (PP (IN on)(NP(NP (DT a) (NN screen))(PP (IN in) (NP (DT a) (JJ darkened) (NN theater))))))))))) | (CC (و) (S (NP (RB بعد) (NP (DTNN الإسلام))) (VP (VBDS أصبحت) (NP (NNP شبام) (NP (NN مدينة) (JJ عامرة))) (EOL .))) |
| **Morphological Analyzer:** | The:1, poing: 31, where: 5, images: 19, were: 2, projected: 29, on: 7, a: 2, screen: 13, in: 10, a: 2, darkened: 2, theatre: 6, .:1 | 11, أصبحت:1, و:1, الاسلام:12, بعد:3, شبام: 1, مدينة:7, عامرة: 8, .:1 |
| **Semantic Analyser** | Point:النقطة, where:التي فيها, images: الصور, were: كانت, projected: عرض, on: على, a: , screen:شاشة, in: في, darkened: مظلم, theater:مسرح | became: أصبحت, after :بعد, و: مدينة: city, عامرة: populated |
| **Translator** | The: ال, .:. | .:. Shibam :شبام , و:, Islam:الاسلام |
| **Replacement** | ال النقطة التي فيها الصور كانت عرض على شاشة في مظلم مسرح Means: The the point where in it the images had viewed on a screen in a theater darkened | After Islam, became Shibam city populated. |
| **Reordering( Target Sentence)** | النقطة التي فيها يتم عرض الصور على شاشة في مسرح مظلم Means: The point where images were projected on a screen in a darkened theater | After Islam, Shibam became a populated city. |

### 3.1.5 Applying the Cosine Similarity

To apply this evaluation we translate each paragraph in the dataset that is accredited by the human translators, using NAN, Google, and Bing. Then we compare the results with the CTLT translations for the same texts. After that, we convert each paragraph to a number of vectors to apply the cosine similarity equation. CTLT translations are also converted. The resulted number is the cosine of an angle that can be found using the $\cos^{-1}$ function.

   ***Step1: converting paragraphs to vectors:***
Each paragraph had a number of words. Those words had a frequent appearance in a single paragraph. For each word, that frequency must be calculated. The words with their frequency are the vectors.

   ***Step2: calculating the cosine similarity function:***

As shown in equation 1, the equation has many fields to be calculated before the final division to get the cosine result. In the numerator, there is a summation of the multiplication for the frequencies of each similar vectors in both compared texts. For each vector that is found in one of them without the other, its frequency must be multiplied by zero. The denominator is calculated by the multiplication of both vectors lengths. These lengths can be calculated using the square root of the power of two for each vector frequency. The cosine similarity function is the division of the numerator by the denominator.

### Step3: calculating the angles of the cosines:

The result of that division will result in a number that is a cosine for a specific angle. If the cosine similarity result is 1, then the angle is 0, which means the full similarity. Applying the inverse of the cosine function, $\cos^{-1}$, to the cosine similarity result will give us that angle. If the resulted angle is small, it is closer to the human translation and vice versa. The results that we get from the 16 paragraphs of the dataset is shown in tables 1 and 2, and in figure 2.

In this way, the model is going to serve the non-native speakers of the English or the Arabic language and therefore they can have a somewhat correct translation compared with the available instant translators. In addition, it will decrease the number of translation errors according to its dependability on the ontology files. NAN-MT, the simulated implementation, is shown in figure 2 below. The f-measure of NAN can be used to evaluate its performance according to equation 3.

$$Precision = \frac{\# of\ correctly\ translated\ words}{\# of\ words\ generated\ by\ the\ chosen\ MT}$$

$$Recall = \frac{\# of\ correctly\ translated\ words}{\# of\ words\ in\ the\ CTLT\ translation}$$

$$F\text{-measure} = \frac{2 * Precision * Recall}{Precision + Recall}$$

Equation 3: F-measure equations

Table 2: F-measure results for NAN, google and Bing

|  | NAN | Google | Bing |
|---|---|---|---|
| precision | 0.949875 | 0.776370 | 0.715439 |
| Recall | 0.996121 | 0.759560 | 0.695829 |
| f-measure | 0.972448 | 0.767873 | 0.705498 |

## 4. THE MODEL EVALUATION AND DISCUSSION

NAN translation has been tested using the precision, recall and f-measure, and so do the translation of Google and Bing. NAN shows values than better than those of Google and Bing. NAN translation in f-measure was approximately to 0.972, where Google in f-measure was about 0.767 and Bing was about 0.705. Those values were got from applying the equations shown in equation 3 and the recorded results are shown in table 2.

In addition, Cosine similarity is used to identify the semantic similarity of NAN translation with CTLT translation. NAN translation is similar to CTLT's due to the ontology-based translation and the correction algorithm. As a result of this example, for the English sentence, Google's angle was 25.33°, Bing's angle was 54.40° and NAN's angle was 0°. In addition, for the Arabic sentence, Google's angle was 21.8°, Bing's angle was 27.06° and NAN's angle was 8.54°. The smallest angle means the most similar translation, among the three, which is recorded for NAN on both sides of the translation (Arabic-English, English-Arabic).

As shown in the tables 3 and 4 and figure 3, NAN's angle is more similar to CTLT translation than Google's and Bing's, because its angle is nearer to the angle of CTLT than the angles of Google and Bing, as highlighted using the bold red font. In each row, we pointed to a paragraph, of the dataset 16 paragraphs, according to its contexts. As we mentioned earlier, the context was used to identify the related ontology file. The smallest recorded angles were NAN's. In this way, we prove that NAN provides Arabic-English translations that are better than any other instant translator, although NAN is not as dynamic as Google and cannot translate texts without knowing their contexts which may result in errors.

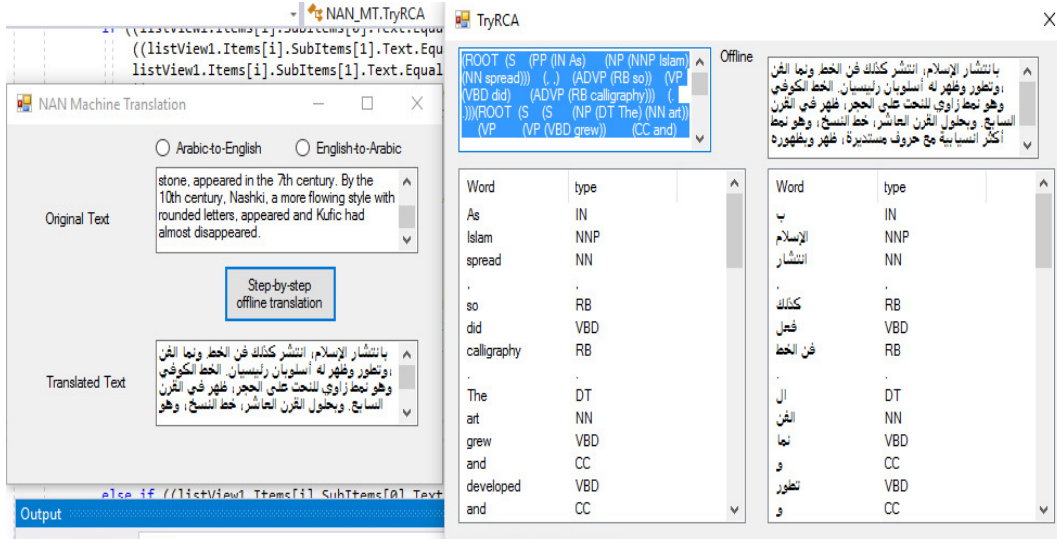

Figure 2: NAN-MT, the simulated implementation

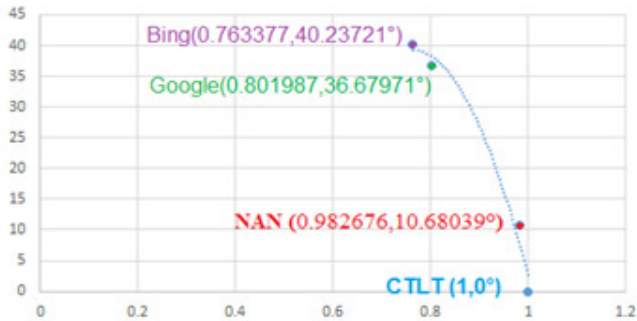

Figure 3: Cosine similarity for the average results

Table 3: The average results of the Cosine Similarity using all the dataset results

| Translator | Cosine | angle |
|------------|--------|-------|
| NAN | 0.982676 | 10.68039° |
| GOOGLE | 0.801987 | 36.67971° |
| BING | 0.763377 | 40.23721° |

The final result for the average cosine and angles is shown in table 3. The detailed results were also recorded in table 4.

## 4. CONCLUSION

The ontology is used to disambiguate the homographs and homonyms which are the main reason for the translation errors. In this paper, we extended a model (NAN) of SWMT for the Arabic-English Translations. The previous model (JM) was ineffective or inefficient if we use it with the Arabic language because of its flexible structure and its complex grammar compared to the English language. Thus, we added additional components to provide somewhat more correct translations than that we got from the original model itself. The available instant translators didn't provide the needed correct translations, too. NAN uses the techniques of JM and the other instant translators.

The cosine similarity proves that NAN translation is better than Google's and Bing's because the resulted angles of NAN is closer to the human translation angles than Google's and Bing's. The results of the evaluation process proved that NAN-MT is better than the other instant translators because it is ontology-based. In a comparison of angles values: the angle of CTLT as the precise translation is equal to 0, NAN's average angle is approximately equal to 10.7°, Google's average angle is approximately equal to 36.6°, and Bing's average angles approximately equal to 40.2°. According to that, NAN translation is approximately more similar to human translation than the other instant translators.

We suppose as a **future work** to rich the ontology files to have more semantically featured translation to make the translation more intelligent. The model is able to work with more complex languages and complex structures. We recommend the university and all other universities to develop a dataset that is applicable for new MTs and those are not sold to the famous Statistical Machine Translation provider such as Google and Bing to provide a new chance to develop the models accurately and in different situations, not just samples.

Table 4: Cosine Similarity for every single paragraph which shows the recorded similarity and their angles

| Paragraph | Google | | Bing | | NAN | |
|-----------|--------|-------|------|-------|-----|-------|
| | Cosine | Angle | Cosine | Angle | Cosine | Angle |
| ShibamE1 | 0.94232 | 19.55512 | 0.899516 | 25.90551 | 0.991744 | 7.367524762 |
| ShibamE2 | 0.934281 | 20.88767 | 0.920575 | 22.98977 | 1 | 0 |
| ShibamA1 | 0.820256 | 34.88959 | 0.768114 | 39.81515 | 0.979809 | 11.53319059 |
| ShibamA2 | 0.819392 | 34.97603 | 0.786067 | 38.18048 | 1 | 0 |
| GlobalizationE1 | 0.911352 | 24.3071 | 0.876734 | 28.74913 | 0.956389 | 16.98353551 |
| GlobalizationE2 | 0.838316 | 33.03729 | 0.578254 | 54.67213 | 1 | 0 |
| GlobalizationA1 | 0.735448 | 42.6549 | 0.620369 | 51.65694 | 0.995166 | 5.636198 |
| GlobalizationA2 | 0.672803 | 47.71622 | 0.590616 | 53.79925 | 0.958241 | 16.61637 |

| CalligraphyE1 | 0.914736 | 23.8317 | 0.853057 | 31.4543 | 1 | 0 |
| CalligraphyE2 | 0.752246 | 41.21471 | 0.805219 | 36.36864 | 0.990783 | 7.785052143 |
| CalligraphyA1 | 0.750234 | 41.38931 | 0.914207 | 23.90657 | 1 | 0 |
| CalligraphyA2 | 0.838398 | 33.02866 | 0.801904 | 36.68767 | 0.991555 | 7.451667975 |
| CinemaE1 | 0.757927 | 40.7182 | 0.796202 | 37.23105 | 0.916694 | 23.55256 |
| CinemaE2 | 0.699933 | 45.57837 | 0.781818 | 38.57265 | 0.965339 | 15.12943 |
| CinemaA1 | 0.726651 | 43.39362 | 0.693296 | 46.10841 | 0.984615 | 10.06327444 |
| CinemaA2 | 0.717501 | 44.15144 | 0.528079 | 58.12427 | 0.992486 | 7.028021906 |

## ACKNOWLEDGMENT

It is our pleasure to thank the Center of Translation and Languages Training (CTLT) of Sana'a University for their collaboration while establishing the sample dataset to test the NAN model.

## DISCLOSURE OF FUNDING AND CONFLICTS

We admit that there are no specific grants from any funding agencies or organizations in the public or even private commercial, or not-for-profit sectors and all was done in a self-payment of the main author.